\documentclass{article}

\usepackage{arxiv}

\usepackage[utf8]{inputenc} 
\usepackage[T1]{fontenc}    
\usepackage{hyperref}       
\usepackage{url}            
\usepackage{booktabs}       
\usepackage{amsfonts}       
\usepackage{nicefrac}       
\usepackage{microtype}      
\usepackage{lipsum}
\usepackage{graphicx}
\usepackage{doi}

\usepackage{tikz}
\usepackage{amsmath}

\usepackage{graphicx} 
\usepackage{tabularx}
\usepackage{subcaption}

\newcommand{\code}[1]{\begin{ttcodefont}#1\end{ttcodefont}}

\usepackage[
backend=biber,
style=numeric,
sorting=ynt
]{biblatex}

\title{Lossless Compression for LLM Tensor Incremental Snapshots}

\author{
{\rm Daniel G. Waddington}\\
IBM Research,\\ 
Almaden Research Center, CA, USA
\and
{\rm Cornel Constantinescu}\\
IBM Research, \\
Almaden Research Center, CA, USA
} 

\date{June 2024}

\begin{document}

\maketitle

\begin{abstract}
During the training of Large Language Models (LLMs), tensor data is periodically "checkpointed" to persistent storage to allow recovery of work done in the event of failure. The volume of data that must be copied during each checkpoint, even when using reduced-precision representations such as \code{bfloat16}, often reaches hundreds of gigabytes.  Furthermore, the data must be moved across a network and written to a storage system before the next epoch occurs.

With a view to ultimately building an optimized checkpointing solution, this paper presents experimental analysis of checkpoint data used to derive a design that maximizes the use of \textit{lossless} compression to reduce the volume of data. We examine how tensor data and its compressibility evolve during model training and evaluate the efficacy of existing common off-the-shelf general purpose compression engines combined with known data optimization techniques such as byte-grouping and incremental delta compression.  

Leveraging our analysis we have built an effective compression solution, known as Language Model Compressor (LMC), which is based on byte-grouping and Huffman encoding. LMC offers more compression performance than the best alternative (BZ2) but with an order-of-magnitude reduction in the time needed to perform the compression.  We show that a 16-core parallel implementation of LMC can attain compression and decompression throughput of 2.78 GiB/s and 3.76 GiB/s respectively. This increase in performance ultimately reduces the CPU resources needed and provides more time to copy the data to the storage system before the next epoch thus allowing for higher-frequency checkpoints.

\end{abstract}

\section{Introduction}

Today, LLMs are typically fine-tuned from existing base models.  To create the base model a compute and data-intensive \textit{pre-training} phase is needed.  This phase uses vast input data sets (e.g., web crawls) to effectively extract language and subject understanding.  For large models, such pre-training can take weeks or even months to complete. For example, the BLOOM 176B parameter model (based on GPT3~\cite{brown2020gpt3}) took 3.5 months to train on a cluster of 384 80G A100 NVIDIA GPUs with a data set of 350B tokens~\cite{workshop2023bloom}.   At this scale of compute, failures invariably occur (\cite{maeng2020cprunderstandingimprovingfailure} report that failures can slow down training by up to 43\%). Prior work has shown that infrastructure and process failures are common in large-scale big data clusters, with a mean time between failure (MTBF) of 4–22 hours~\cite{failures2017,analysisfailure2014}. Consequently, resulting outages and restarts can become very costly.

\begin{figure}
    \centering
    \includegraphics[width=.7\linewidth]{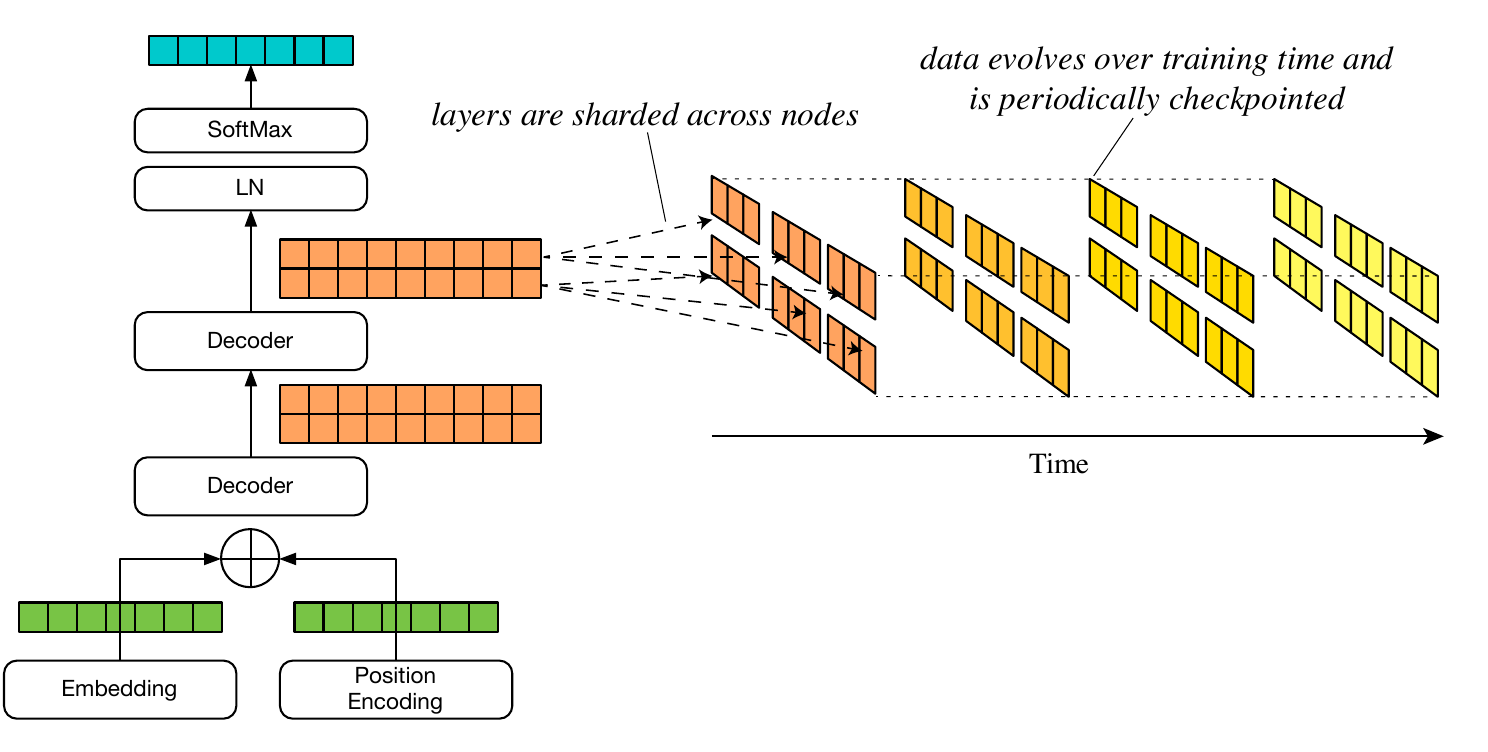}
    \caption{Tensor Data Shards Evolving }
    \label{fig:sharding}
\end{figure}

Scale-out of the training means that the compute and data are distributed across a cluster of compute nodes.  Parallelism is achieved either distributing the training pipeline or by distributing the model weights and biases through \textit{sharding}.  The size of the checkpointed data depends on the model and training scheme.  Nevertheless, a single checkpoint, combining data from all nodes, can reach many gigabytes.  For example BLOOM checkpoints are 329 GiB in size consisting of data sharded across 48 nodes and 384 GPUs~\cite{workshop2023bloom}. As the training occurs the tensors are updated over time (see Figure~\ref{fig:sharding}).

To reduce cost incurred by restarting due to failure in the cluster, \textit{checkpointing} is commonly used. Checkpointing periodically creates a snapshot of GPU memory state at a single point in time.  Checkpoints are coherent across multiple nodes in the compute cluster - this is normally achieved by suspending compute at a specified epoch\footnote{An epoch is one full cycle through the input data}.  Once the state has been copied from the GPU memory (e.g., into local host memory) the training workload can asynchronously continue while the checkpoint data is written out to persistent storage.  

There is a balance between checkpointing frequency, the volume of data to persist, the time the GPUs are waiting idle, and the time to recover after failure.  Prior work focuses on various optimizations of checkpoint frequency, reducing GPU idle  time and minimizing recovery time ~\cite{checkfreq,checknrun,gemini2023}. 

This paper focuses on reducing the volume of data through lossless compression and the resources and time needed to perform it. We focus on lossless compression so as to allow precise recovery and elimination of unexpected possible side-effects that might result from lossy compression.  

We start by first examining data
from six LLM checkpoint sequences taken from Hugging Face (HF).  We examine attributes of the data (type, shape, size, entropy) and how it evolves during the training process. This is followed by an evaluation of the effectiveness of applying common-of-the-shelf (COTS) compression engines. From an understanding of the data, in Section~\ref{sec:proposed}, we propose an alternative compression scheme based on Byte-Grouping (BG), Huffman Encoding and Run-Length Encoding (RLE) that offers more compressability and reduced execution time than alternative COTS encoders.  We evaluate our method and compare it with the best known alternatives.

\section{Background: The Nature of LLM Tensors}

\begin{figure}
    \centering
    \includegraphics[width=0.8\linewidth]{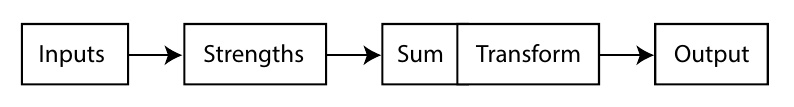}
    \caption{Functional description of a biological neuron’s structure~\cite{fundamentals2017}}
    \label{fig:neuronfunction}
\end{figure}

Deep learning is a form of machine learning that builds large neural networks similar in essence to the human brain~\cite{fundamentals2017}.  The neural network is composed of set of interconnected layers whereby artificial neurons "activate" according to an operation on their inputs.  For dense networks, each neuron is connected to every other neuron of the next layer, hence its output value becomes the input for the next layer. During training, there are two neuron parameters that are adjusted: \textit{weight} and \textit{bias}.  Weights are coefficients that affect the "strength" of each input \((w_ij \cdot x_i)\), that is, how much of the input data is fed into the activation function (see Figure~\ref{fig:neuronfunction}). The weighted inputs are then summed and transformed using an activation function \(f\) to generate the output \(y_j\) as follows:

\[y_j = f(\sum(w_ij \cdot x_i) + b_j)\]

During the training phase of a neural network, weights and biases are adjusted iteratively to minimize the difference between the network’s predictions and the desired outcome.  While LLM models and training is a complex matter outside the scope of this paper, these deep learning fundamentals carry through.

Neuron parameters are represented as numerical \textit{tensors}.  The term tensor is a generalisation of scalars and vectors. Tensor elements are floating point during training and often quantized to integer values for the purpose of inferencing.  



\begin{figure}
    \centering
    \includegraphics[width=1\linewidth]{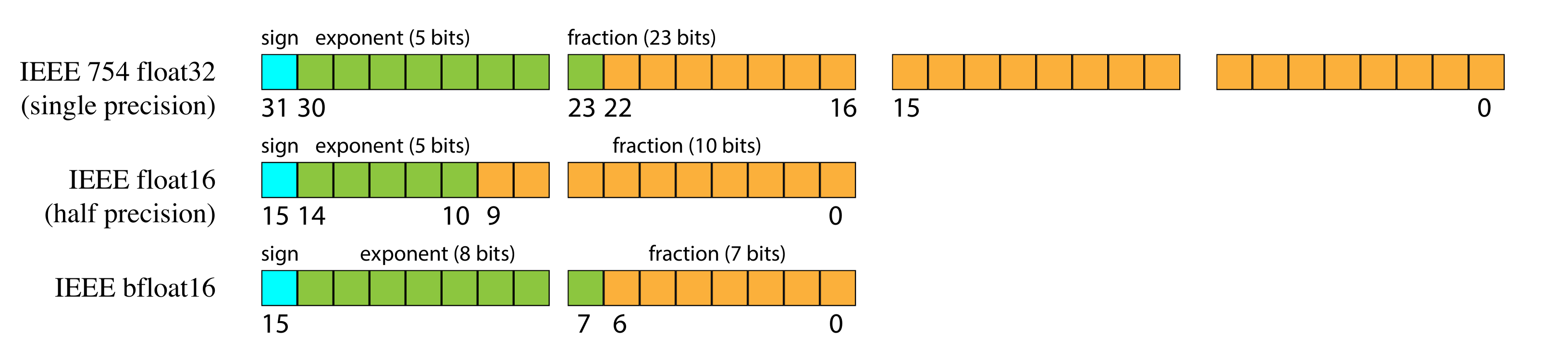}
    \caption{LLM Floating Point Formats}
    \label{fig:floatformats}
\end{figure}

\paragraph{Floating-Point Representations}
There are many bit representations of floating point numbers available in today's computer systems.  Most commonly, LLM models are based on one of three types: IEEE 32-bit float (single precision), IEEE 16-bit float (half precision) and Google's 16-bit brain float~\cite{bfloat2019} (see Figure~\ref{fig:floatformats}).  These formats differ by how many bits are available and how the bits are divided between the exponent and mantissa (also termed the fraction).  Because most hardware now supports bfloat16 as a superior format for deep learning (performing nearly as well as float32 but using half the number of bits)\cite{bfloat2019}, this paper focuses on the analysis of checkpointing data in float32 and bfloat16 formats (see Table~\ref{tab:samples_spec}).

\section{Analysis of LLM Checkpoint Data}

We examine checkpoint data for six different models that are all publicly available on Hugging Face (\code{https://huggingface.co/}). From these sources, we have analyzed over 28 TiB of tensor data.  Three of the data sets are based on bfloat16 and three on float32. The attributes of the sample models are shown in Table~\ref{tab:samples_spec}.   Additional information about the sample data is provided in Appendix A.

\begin{table}[!t]

\begin{centering}
\begin{tabularx}{1.0\linewidth}{
>{\setlength{\hsize}{.25\hsize}\raggedright\footnotesize}X
>{\setlength{\hsize}{.15\hsize}\raggedright\footnotesize}X
>{\setlength{\hsize}{.10\hsize}\raggedright\footnotesize}X
>{\setlength{\hsize}{.1\hsize}\raggedright\footnotesize}X
>{\setlength{\hsize}{.15\hsize}\raggedright\footnotesize}X
>{\setlength{\hsize}{.1\hsize}\raggedright\footnotesize}X
>{\setlength{\hsize}{.1\hsize}\raggedright\arraybackslash\footnotesize}X }
  \hline
  \textbf{Name} & \textbf{Model Size\ (params)}  & \textbf{Tokens} & \textbf{Type} & \textbf{Chkpt Size} & \textbf{Chkpt Count} & \textbf{Shard Splits}\\
  \hline
  bloom       & 176B  & 366B  & BF16 & 329GiB & 20  & 72 \\
  amber       & 6.74B & 1.25T & BF16 & 13GiB  & 358 & 3 \\
  transnormer & 7B    & 1.4T  & BF16 & 28GiB  & 18  & 11 \\
  multiberts  & 110M  &  160M & FP32 & 441MiB & 28 & 1 \\
  olmo        & 7B    & 2.5T  & FP32 & 26GiB  & 558 & 1 \\
  gpt2    & 124M  & 4M    & FP32 & 1.4GiB & 2000 & 1 \\
  \hline
\end{tabularx}
\caption{Attributes of Analyzed Data Sets}
\label{tab:samples_spec}
\end{centering}
\end{table}

\subsection{Converging Behavior}

\begin{figure}
    \centering
    \includegraphics[width=0.7\linewidth]{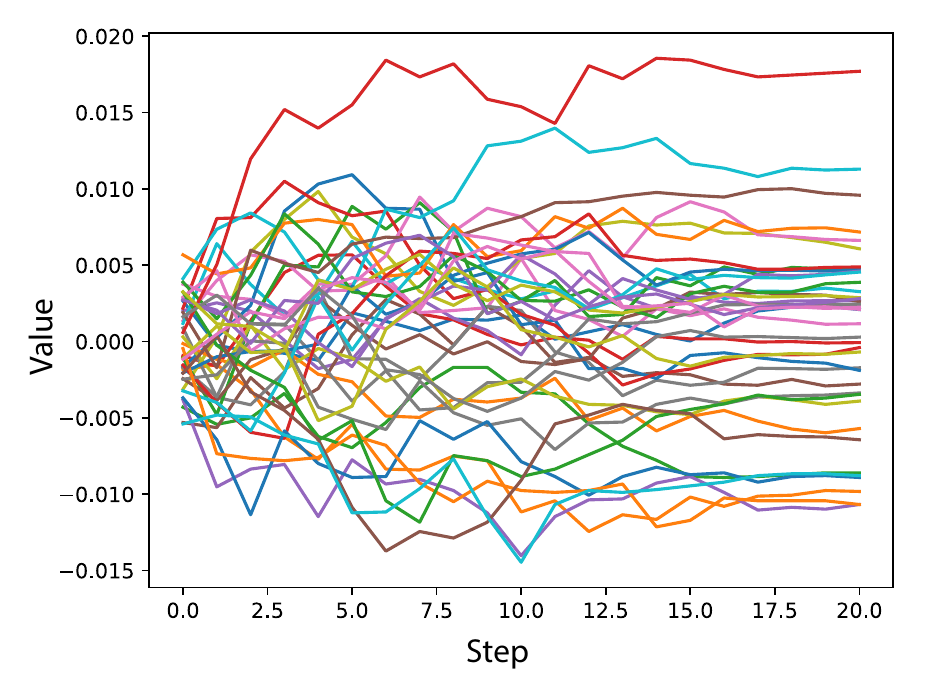}
    \caption{Sample Self-Attention Weights over Time from BLOOM Data Set}
    \label{fig:bloom-sample-weights}
\end{figure}

As previously discussed, the purpose of training a model is to "tweak" the parameters until the optimum function of the neural network can be achieved.  This adjustment of parameters lessens over time, eventually reaching values that are stable.  Figure~\ref{fig:bloom-sample-weights} shows a sample of weights (from shard \code{h.18.self\_attention.dense.weight}) for the BLOOM data set. As is evident from the plot, at step 0 the values start randomly within the range ~+/- 0.005, then after a period of relatively aggressive change, they start to converge by around step 17. By step 20, the values are steady. This convergence behavior indicates that there are potential gains available for \textit{delta encoding} between checkpoints.

\begin{figure}
    \centering
    \includegraphics[width=0.7\linewidth]{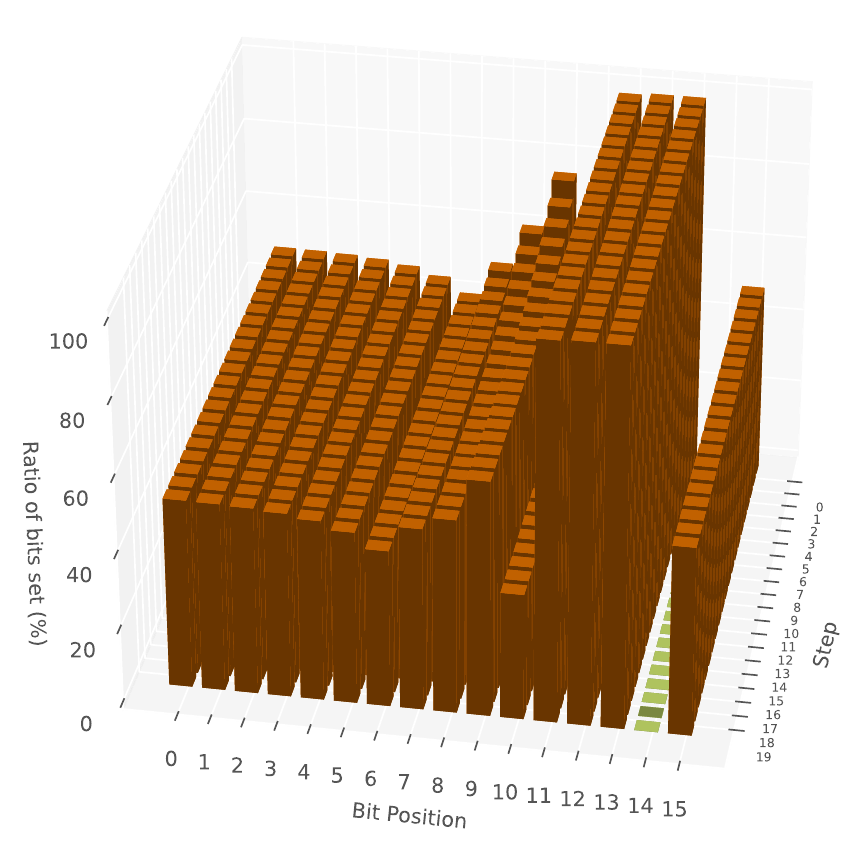}
    \caption{Bit Distribution for BLOOM Self-Attention QKV Weights Shard 0}
    \label{fig:bloom-bit-distro}
\end{figure}

We now examine the bit distribution, that is the ratio of zeros and ones.
 This data is from BLOOM and is therefore bfloat16.  Figure~\ref{fig:bloom-bit-distro} shows what proportion of bits (as a percentage on the z-axis) are set at that single point in time.  Note that step 0 is at the rear of the plot, i.e., time runs from the back to the front.

Although the early steps are more volatile, the plot shows that most bits (except 10 and 14) flip between steps. Bit 14 is the most significant bit of the exponent (bit 15 is the sign bit) and is thus only set for values beyond the range of 7-bit exponent (\(3.4*10^{19}, -3.4*10^{19} = 6.4*10^{19}\)). In this data, the values are not reaching a magnitude to warrant this bit.

To explore further how tensor data changes over time, we examine how many bits are flipped between "steps" (refer to Figure~\ref{fig:sharding}), which represent an iteration of the whole input data set.  We do this by examining the result of an XOR operation on values in step $n$ and $n+1$.

\begin{figure}[ht!]
    \centering
    \includegraphics[width=0.7\linewidth]{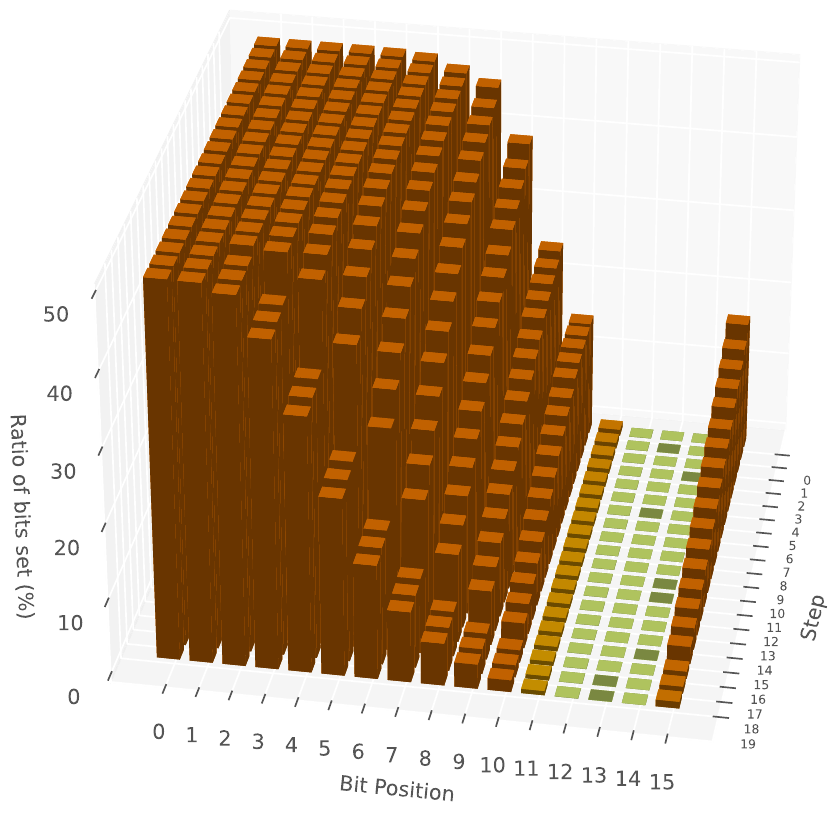}
    \caption{XOR Delta Bit Distribution for BLOOM Self-Attention QKV Weights Shard 0}
    \label{fig:bloom-bit-xor-distro}
\end{figure}

Figure~\ref{fig:bloom-bit-xor-distro} shows the change in bit flip ratios as the step number increases.  Note that step 0 is furthest back on the y-axis and that the z-axis maximum is 50\% (i.e., equivalent to random change).  This plot confirms several things.  First, as indicated by falling bit set ratios (z-axis), stability increases as steps progress.  Second, least-significant bits change more than most-significant bits - bits 10-14 rarely change at all.  Third, sign bit changes reduce as the learning progresses, indicating less values are fluctuating across the zero boundary. In all, the "space available" in the three-dimensional cube suggest potential for compression. 

\begin{figure}
    \centering
    \includegraphics[width=0.5\linewidth]{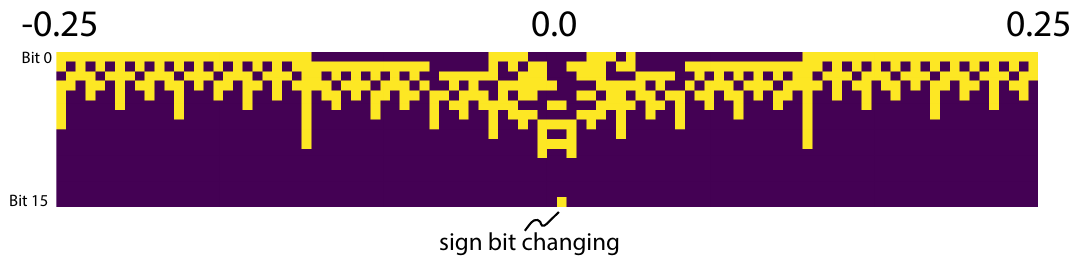}
    \caption{Bit changes for bfloat16 values -0.25 to 0.25 modified in 0.005 increments}
    \label{fig:bfloat16-bit-changes}
\end{figure}

While, Figure~\ref{fig:bloom-bit-xor-distro} shows increasing potential for compressability as learning progresses, it is clear that the lowest-significant bits (especially bits 0-4) continue to fluctuate even close to convergence.  This is congruent with the nature of floating point representations.  Figure~\ref{fig:bfloat16-bit-changes} illustrates how bfloat16 bits change as 0.005 increments are made from -0.25 to 0.25.  The yellow pixels indicate a bit flip (between the value and its incremented value). Bit 0 is at the top of the figure.  From this figure, it is evident that even for very small increments, the first few bits almost always change.


    

\section{Proposed Compression Algorithm}
\label{sec:proposed}
\paragraph{Byte-Grouping and RLE} \label{byte-grouping}

\begin{figure}
    \centering
    \includegraphics[width=0.5\linewidth]{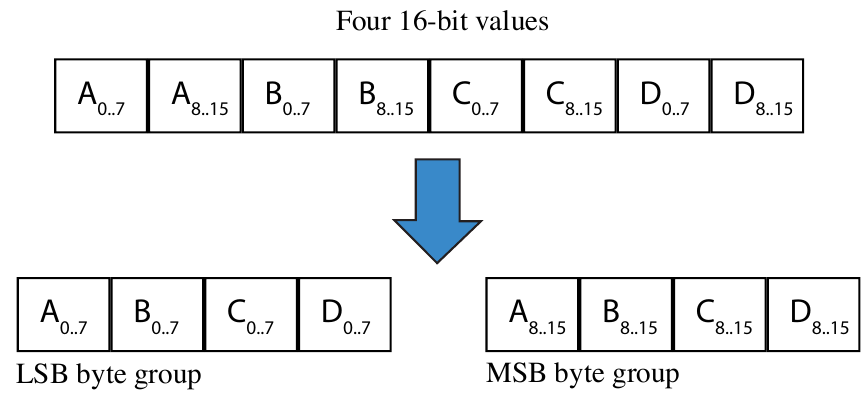}
    \caption{Example Byte-Grouping of four 16-bit values}
    \label{fig:bytegroup}
\end{figure}

Byte-grouping (BG) is a simple modelling strategy that rearranges the data to exploit different stability across bytes.  For example, considering Figure~\ref{fig:bloom-bit-xor-distro}, the stability of the most-significant byte (bits 8-15) is much greater than the least-significant byte (bits 0-7). Byte-grouping copies bytes from their original position in the tensor value into contiguous regions of memory assigned to the group.  For 16-bit values, two byte-groups are formed (see Figure~\ref{fig:bytegroup} example). For 32-bit values, four byte-groups are formed.

Grouping more stable bytes in contiguous regions of memory, leads to more effective Run Length Encoding (RLE).  RLE is a well known encoding method that replaces a symbol $S$ with the pair $(S,c)$ where $c$ is the number of times $S$ is repeated in this position.  For example:
\[
    AAAABBBAACCC \rightarrow (A,4),(B,3),(A,2),(C,3)
\]

Thus, RLE is effective when symbols are repeated. For the tensor delta data, repeating symbol zero is common and therefore RLE offers a path to additional compression.

\subsection{Estimating compressibility with Entropy}

Byte-grouping transformation seems to help many off-the-shelf compression engines so we are using this layout of data. Here we want to investigate what would be the compressibility of this type of data when using a simple model: considering bytes as independent compression symbols and measuring per symbol (byte) information content. The average amount of information per symbol over all alphabet (of 256 symbols) is known as the \textit{entropy} of the symbols probability distribution and measured in bits is given by:
\[
H = - \sum_{i=0}^{255} p_i \log_2 p_i
\]
where $p_i$ denotes the probability of occurrence of symbol $i$ (byte value $i$) of the alphabet. For a given probability distribution of the symbols, the entropy represents the minimum number of bits needed to represent a symbol without losing any information (i.e. the code to be uniquely decodable). Note that the worst case for the entropy is when the symbols are uniformly distributed, so the probability of each symbol is $p_i = 1/256$ and the entropy $H = 8$ bits.

\begin{figure}
    \centering
    \includegraphics[width=.7\linewidth]{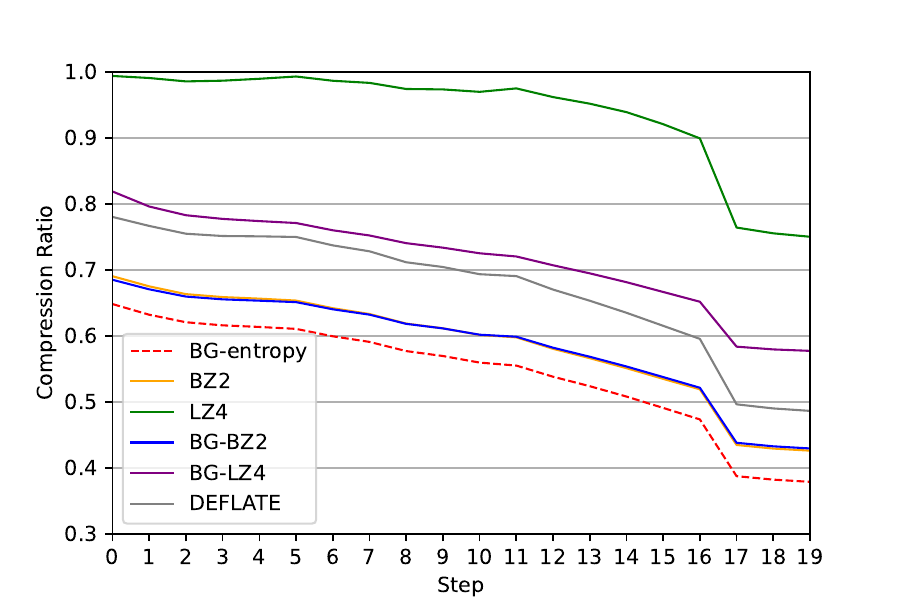}
    \caption{Entropy Comparison with Off-the-shelf Engines on BLOOM Data}    
    \label{fig:entropy-vs-offtheshelf}
\end{figure}

To capture the changes in symbols probability distribution we provide \textit{adaptivity} by computing symbols probability distribution on relatively small blocks of maximum $64KiB$ (instead of entire file). As symbol probability $p_i$ we use:
\[
p_i = \frac{\text{number of occurrences of the symbol $i$ in the block}}{\text{block size in bytes}}
\]
Figure~\ref{fig:entropy-vs-offtheshelf} compares the compression ratios achieved by BZ2, DEFLATE and LZ4 (common off-the-shelf) with that which could be achieved by an entropy-based compression approach. BZ2 compression algorithm comes closer to the entropy, but it involves complex modelling by preprocessing data using the Burrows-Wheeler transform~\cite{BWT} making it quite slow. 

Next, we examine an alternative entropy coder implementation that achieves the code size per symbol (byte) closer to entropy.


\subsection{Large Model Compression (LMC)}

\begin{figure}
    \centering
    \includegraphics[width=.7 \linewidth]{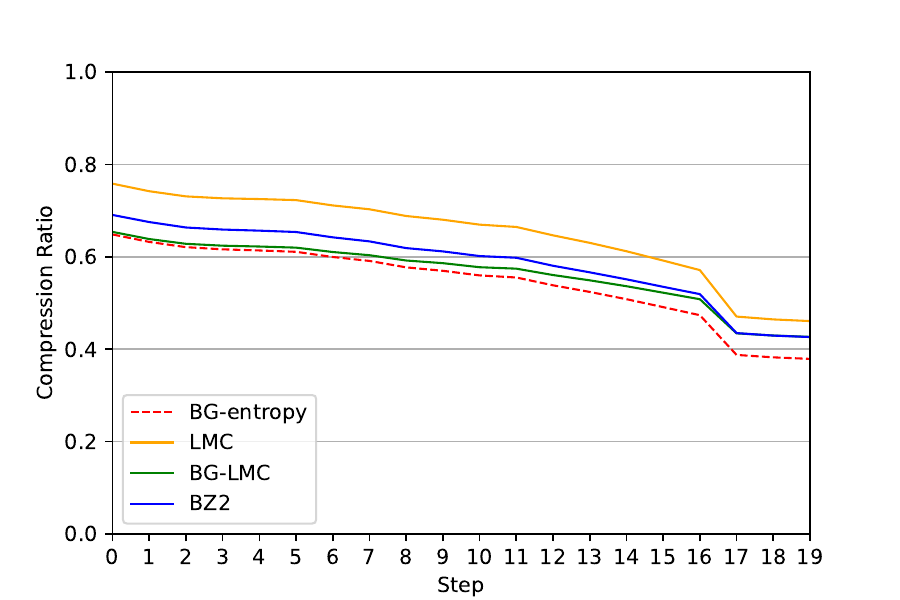}
    \caption{Comparison of BZ2, LMC, BG-LMC compression ratios vs. entropy of BLOOM data}
    \label{fig:entropy-LMC}
\end{figure}

We investigate using a Huffman-based entropy encoder because it is fast and optimal within 1 bit of entropy. A somewhat better alternative (in terms of compression ratio) would be an arithmetic coder.  However, arithmetic encoding is generally much slower than Huffman encoding. To adapt to changes in per byte statistics as for entropy estimation, we eventually compute a new Huffman codebook for each block of maximum $64KB$ and encode it in compressed stream (similar to~\cite{rfc1951},~\cite{Hirsch1990}). Additionally, if the entire block consists of one byte value  $S$  repeated, it is encoded as a pair $(S,c)$, as in paragraph~\ref{byte-grouping}. In short, LMC compression scheme can be summarized as block adaptive Huffman coding + RLE.
Figure~\ref{fig:entropy-LMC} compares BG-LMC with BG-entropy showing that BG-LMC comes very close to the ideal compression ratio offered by the entropy.  The data shows that as the input becomes more compressible, the entropy ratio becomes more difficult to achieve due to the use of integral bits/byte inherent to Huffman coding. 


\subsection{Parallel LMC (PLMC)}

We implemented a simple data-parallel version of the BG-LMC algorithm (referred to as PLMC) that iteratively reads  byte-grouped data into an input buffer and uses OpenMP to implement multi-threaded compression and decompression.  The number of segments corresponds to the number of threads and is included in the codestream metadata, together with the compressed size of each piece. The implementation uses a default buffer size of 128 MiB.  Nevertheless, this can be configured.

\section{Experimental Results}

The system used to collect the data is based on an AMD Epyc 7343 16-core processor with 512GiB main memory and 23TiB NVMe SSD.

\subsection{Single-threaded Performance}
\paragraph{Compression Ratios and Processing Time}

We collected data for compression ratio, compression throughput and decompression throughput. To better abstract the data, we calculated mean values across shard-deltas belonging to each data source.  The primary attributes for
consideration are compression ratio and throughput.  For our use-case, compression performance is more important than decompression performance because the latter only happens in the event of disaster-recovery.

\begin{figure}
    \centering
    \includegraphics[width=.8\linewidth]{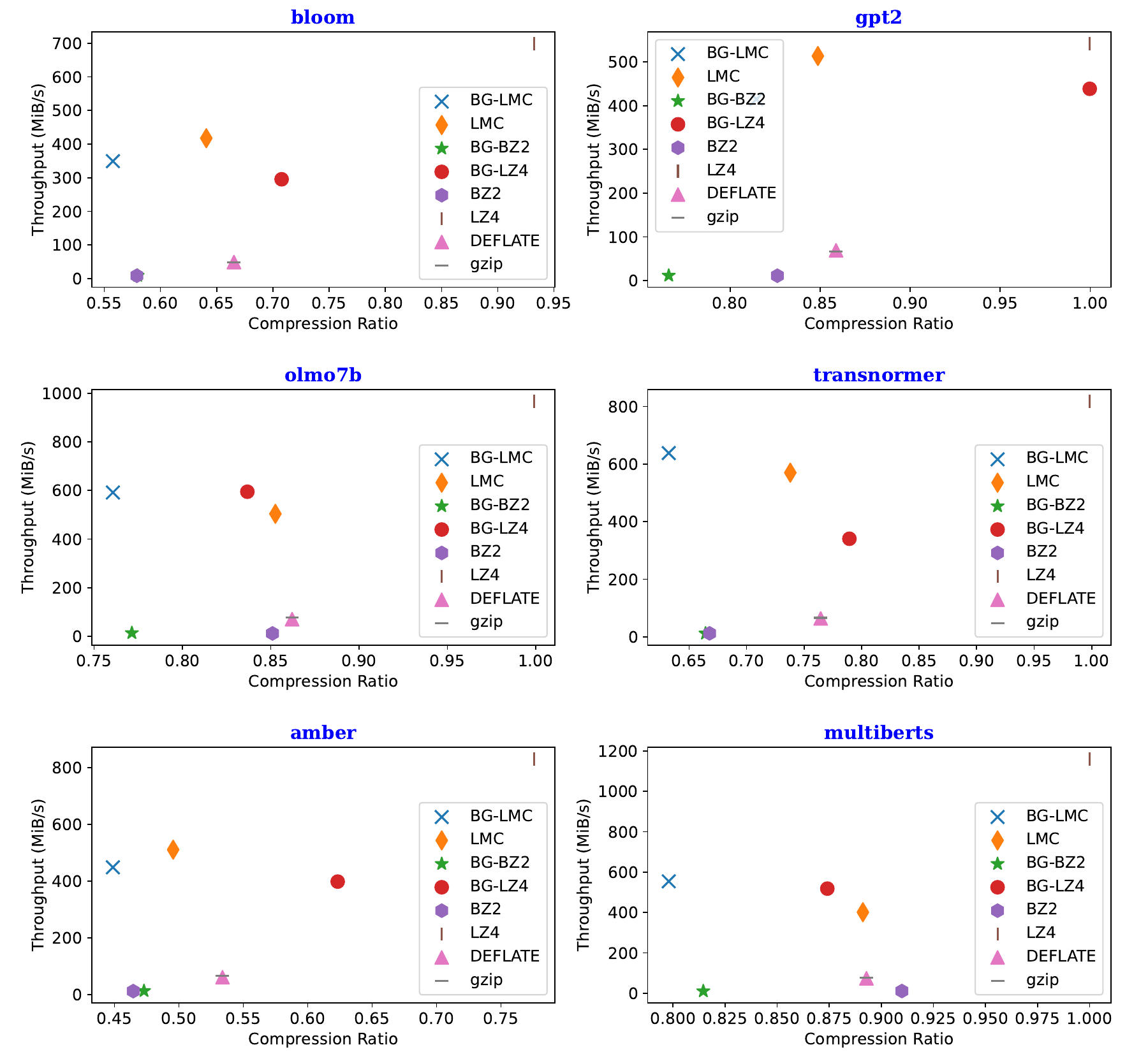}
    \caption{Summary of Encoder Performance}
    \label{fig:7plot-ratio-time}
\end{figure}

Figure~\ref{fig:7plot-ratio-time} plots mean compression ratio vs. compression throughput for each of the six compression engines.  The plots indicate that BG-LMC performs consistently better than the others. While BG-BZ2 achieves compression ratios within a few percent of BG-LMC its compression throughput performance is an order of magnitude slower.  On the other end of the spectrum, the data shows that LZ4 is typically the fastest to compress, but offers a relatively poor compression ratio - the best achieved being only 0.77 for the \textbf{amber} data set.

Appendix 3 shows data for decompression for a sample of five shards.  BG-LMC decompression is 1.4x slower than the fastest decompression (LZ4).





\paragraph{Compression Variance Over Time}

We next examine how the compressability changes over time.  As previously discussed, as convergence occurs the tensor values stabilize and thus the delta changes reduce - resulting in more zeros and therefore more compressability.

\begin{figure}[ht!]
    \centering
    \includegraphics[width=.8\linewidth]{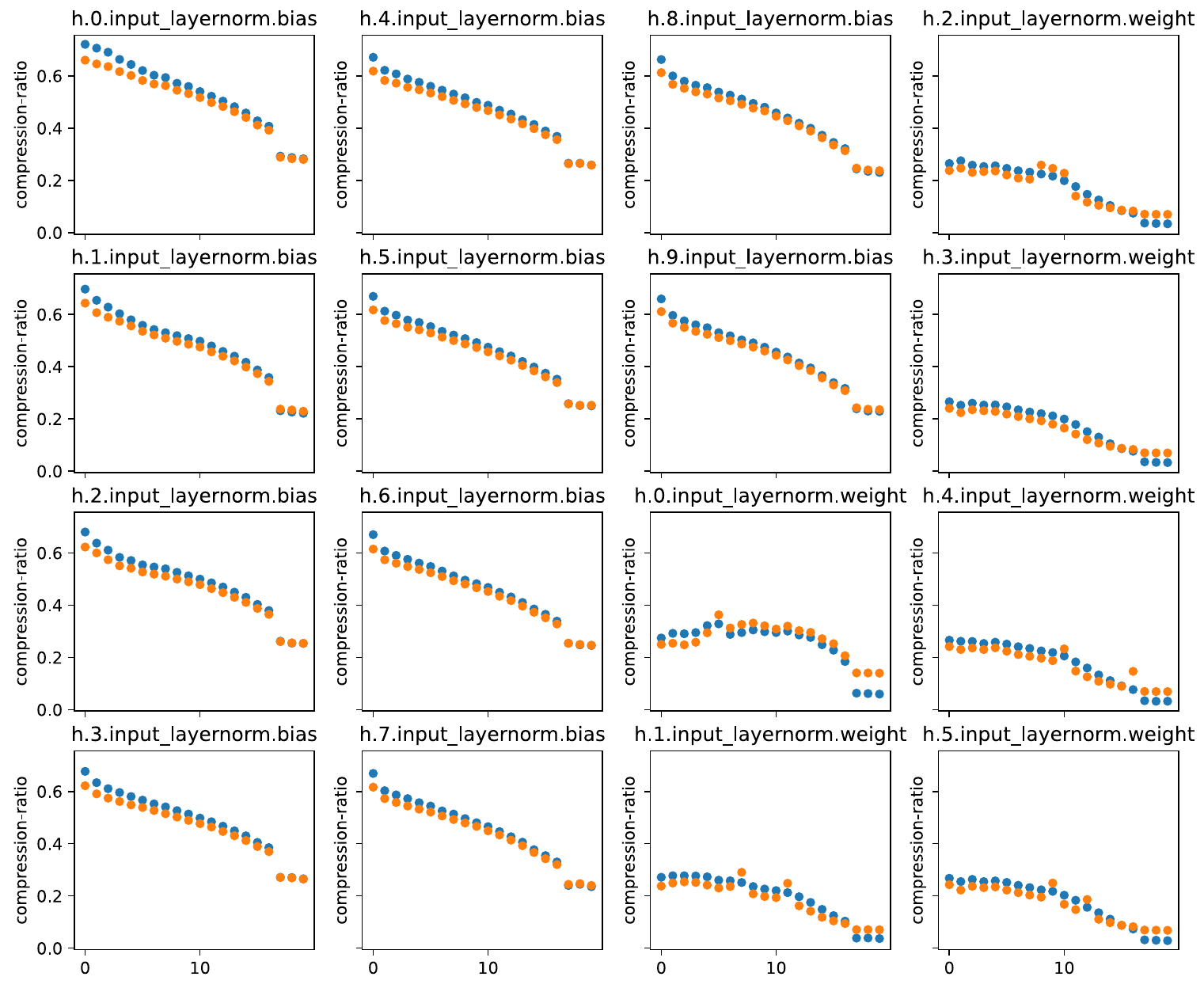}
    \caption{Compressability over time for 16 BLOOM shards\\Orange:BG-LMC, Blue:BZ2}
    \label{fig:compression-over-time-sample}
\end{figure}

Figure~\ref{fig:compression-over-time-sample} shows how compressibility of shard-deltas changes over time for 16 samples (mixed weights and biases). We show data for BG-LMC and BZ2.  In this sample, the compression ratio for biases changes from $\sim$0.7 to $\sim$0.2 and for weights changes from $\sim$0.4 to less than 0.1. As the data converges, the compressibility gradient flattens out.

\subsection{Multi-core Scale-up Performance}

\begin{figure}[!h]
    \centering
    \includegraphics[width=0.5\linewidth]{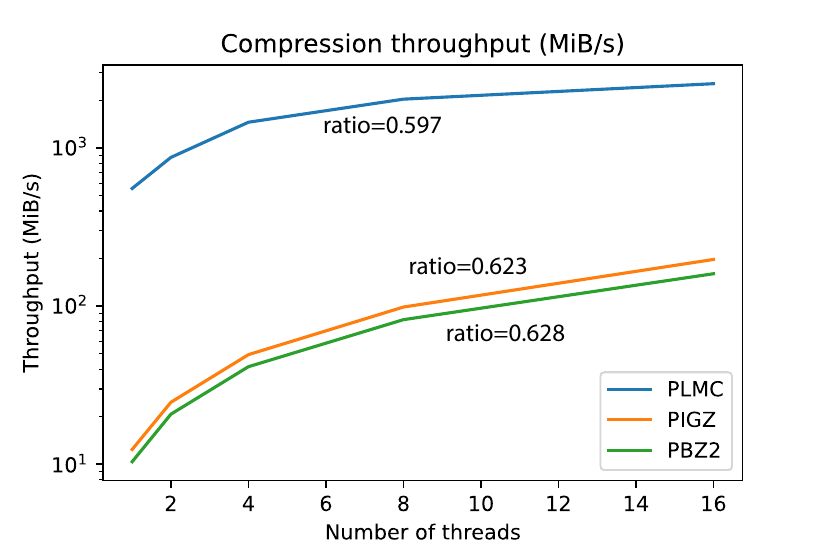}
    \caption{Comparison of PBZ2, PIGZ, PLMC compression throughput}
    \label{fig:comp-throughput-PLMC}
\end{figure}

\begin{figure}[!h]
    \centering
    \includegraphics[width=0.5\linewidth]{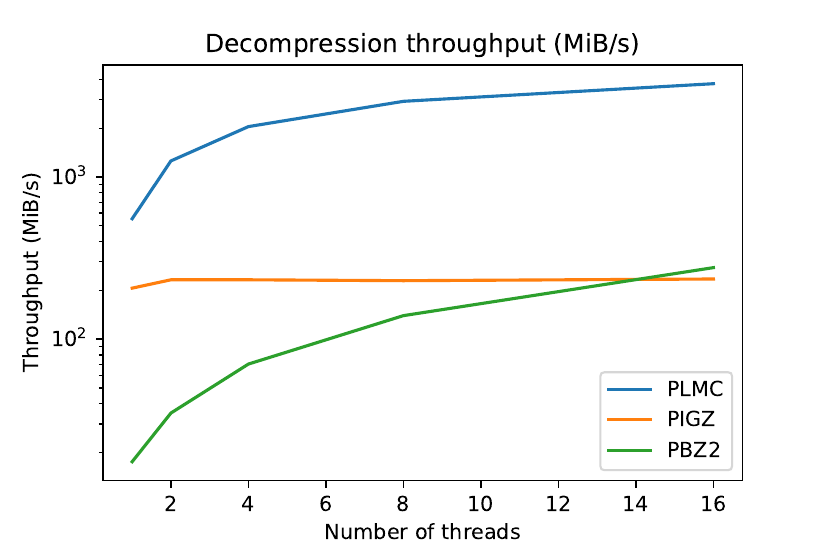}
    \caption{Comparison of PBZ2, PIGZ, PLMC decompression throughput (MiB/s)}
    \label{fig:decomp-throughput-PLMC}
\end{figure}

To evaluate throughput scale-up, we evaluated concurrent implementations of LMC, GZIP and BZ2 - all on top of byte-grouped (BG) data.  For these measurements IO bottlenecks were greatly reduced by using a RAM-disk.  We chose a buffer size of 128 MiB although we found that throughput is not sensitive to buffer size.  Figure~\ref{fig:comp-throughput-PLMC} compares the compression throughput for the PLMC and off-the-shelf parallel implementations PIGZ and PBZ2. Figure~\ref{fig:decomp-throughput-PLMC} compares the decompression throughput. The compression ratio achieved by PLMC is $0.597$ exceeding PIGZ ($0.623$) and PBZ2 ($0.628$) but with a significantly higher throughput.  The data shows that compression scales relatively well for all three encoders.  PIGZ decompression does not scale well.

\section{Related Work}
As a general survey ~\cite{park2024comprehensivesurveycompressionalgorithms} presents compression algorithms for language models, including lossy-based solutions such as quantization and parameter sharing. Although LLM tensor data is not covered, ~\cite{10.14778/3648160.3648180} presents analysis and benchmarking of lossless compression for floating-point data. 

Prior work~\cite{hershcovitch2024losslessnearlosslesscompressionfoundation} has also explored the use of byte-grouping for tensor data for storage purposes. They do not cover incremental deltas nor improve on existing compression engines. Finally, ~\cite{10.1145/3560814} presents a general survey of time-series compression including work on combining Huffman encoding with delta and RLE~\cite{8546121}.


\newpage
\section{Conclusions}

In this paper we have explored the compressability of LLM tensor data for incremental snapshotting to storage.  We examine how the tensor-deltas evolve over time down, how they are represented and how individual bits behave.  With this knowledge in hand, we determined that entropy-based compression using Huffman encoding offers a superior solution to well known generic alternatives such as BZ2 and LZ4.

Our solution, BG-LMC, combines fast-modelling (byte-grouping and RLE) with Huffman encoding to achieve compression levels marginally better than the best off-the-shelf alternative (BZ2) but with an order of magnitude reduction in time-to-compress.  This analysis shows using BG-LMC can achieve compression ratios of less than 0.6 and single-threaded compression throughput of $\sim$300 MB/s.  For multi-core scale-up, our solution achieves performance beyond parallel implementations of gzip and BZ2, attaining over 2.5GiB/s and 3.6GiB/s for compression and decompression respectively on only 16 cores.

\vspace{3cm}
\printbibliography

@misc{workshop2023bloom,
      title={BLOOM: A 176B-Parameter Open-Access Multilingual Language Model}, 
      author={BigScience Workshop and : and Teven Le Scao and Angela Fan and Christopher Akiki and Ellie Pavlick and Suzana Ilić and Daniel Hesslow and Roman Castagné and Alexandra Sasha Luccioni and François Yvon and Matthias Gallé and Jonathan Tow and Alexander M. Rush and Stella Biderman and Albert Webson and Pawan Sasanka Ammanamanchi and Thomas Wang and Benoît Sagot and Niklas Muennighoff and Albert Villanova del Moral and Olatunji Ruwase and Rachel Bawden and Stas Bekman and Angelina McMillan-Major and Iz Beltagy and Huu Nguyen and Lucile Saulnier and Samson Tan and Pedro Ortiz Suarez and Victor Sanh and Hugo Laurençon and Yacine Jernite and Julien Launay and Margaret Mitchell and Colin Raffel and Aaron Gokaslan and Adi Simhi and Aitor Soroa and Alham Fikri Aji and Amit Alfassy and Anna Rogers and Ariel Kreisberg Nitzav and Canwen Xu and Chenghao Mou and Chris Emezue and Christopher Klamm and Colin Leong and Daniel van Strien and David Ifeoluwa Adelani and Dragomir Radev and Eduardo González Ponferrada and Efrat Levkovizh and Ethan Kim and Eyal Bar Natan and Francesco De Toni and Gérard Dupont and Germán Kruszewski and Giada Pistilli and Hady Elsahar and Hamza Benyamina and Hieu Tran and Ian Yu and Idris Abdulmumin and Isaac Johnson and Itziar Gonzalez-Dios and Javier de la Rosa and Jenny Chim and Jesse Dodge and Jian Zhu and Jonathan Chang and Jörg Frohberg and Joseph Tobing and Joydeep Bhattacharjee and Khalid Almubarak and Kimbo Chen and Kyle Lo and Leandro Von Werra and Leon Weber and Long Phan and Loubna Ben allal and Ludovic Tanguy and Manan Dey and Manuel Romero Muñoz and Maraim Masoud and María Grandury and Mario Šaško and Max Huang and Maximin Coavoux and Mayank Singh and Mike Tian-Jian Jiang and Minh Chien Vu and Mohammad A. Jauhar and Mustafa Ghaleb and Nishant Subramani and Nora Kassner and Nurulaqilla Khamis and Olivier Nguyen and Omar Espejel and Ona de Gibert and Paulo Villegas and Peter Henderson and Pierre Colombo and Priscilla Amuok and Quentin Lhoest and Rheza Harliman and Rishi Bommasani and Roberto Luis López and Rui Ribeiro and Salomey Osei and Sampo Pyysalo and Sebastian Nagel and Shamik Bose and Shamsuddeen Hassan Muhammad and Shanya Sharma and Shayne Longpre and Somaieh Nikpoor and Stanislav Silberberg and Suhas Pai and Sydney Zink and Tiago Timponi Torrent and Timo Schick and Tristan Thrush and Valentin Danchev and Vassilina Nikoulina and Veronika Laippala and Violette Lepercq and Vrinda Prabhu and Zaid Alyafeai and Zeerak Talat and Arun Raja and Benjamin Heinzerling and Chenglei Si and Davut Emre Taşar and Elizabeth Salesky and Sabrina J. Mielke and Wilson Y. Lee and Abheesht Sharma and Andrea Santilli and Antoine Chaffin and Arnaud Stiegler and Debajyoti Datta and Eliza Szczechla and Gunjan Chhablani and Han Wang and Harshit Pandey and Hendrik Strobelt and Jason Alan Fries and Jos Rozen and Leo Gao and Lintang Sutawika and M Saiful Bari and Maged S. Al-shaibani and Matteo Manica and Nihal Nayak and Ryan Teehan and Samuel Albanie and Sheng Shen and Srulik Ben-David and Stephen H. Bach and Taewoon Kim and Tali Bers and Thibault Fevry and Trishala Neeraj and Urmish Thakker and Vikas Raunak and Xiangru Tang and Zheng-Xin Yong and Zhiqing Sun and Shaked Brody and Yallow Uri and Hadar Tojarieh and Adam Roberts and Hyung Won Chung and Jaesung Tae and Jason Phang and Ofir Press and Conglong Li and Deepak Narayanan and Hatim Bourfoune and Jared Casper and Jeff Rasley and Max Ryabinin and Mayank Mishra and Minjia Zhang and Mohammad Shoeybi and Myriam Peyrounette and Nicolas Patry and Nouamane Tazi and Omar Sanseviero and Patrick von Platen and Pierre Cornette and Pierre François Lavallée and Rémi Lacroix and Samyam Rajbhandari and Sanchit Gandhi and Shaden Smith and Stéphane Requena and Suraj Patil and Tim Dettmers and Ahmed Baruwa and Amanpreet Singh and Anastasia Cheveleva and Anne-Laure Ligozat and Arjun Subramonian and Aurélie Névéol and Charles Lovering and Dan Garrette and Deepak Tunuguntla and Ehud Reiter and Ekaterina Taktasheva and Ekaterina Voloshina and Eli Bogdanov and Genta Indra Winata and Hailey Schoelkopf and Jan-Christoph Kalo and Jekaterina Novikova and Jessica Zosa Forde and Jordan Clive and Jungo Kasai and Ken Kawamura and Liam Hazan and Marine Carpuat and Miruna Clinciu and Najoung Kim and Newton Cheng and Oleg Serikov and Omer Antverg and Oskar van der Wal and Rui Zhang and Ruochen Zhang and Sebastian Gehrmann and Shachar Mirkin and Shani Pais and Tatiana Shavrina and Thomas Scialom and Tian Yun and Tomasz Limisiewicz and Verena Rieser and Vitaly Protasov and Vladislav Mikhailov and Yada Pruksachatkun and Yonatan Belinkov and Zachary Bamberger and Zdeněk Kasner and Alice Rueda and Amanda Pestana and Amir Feizpour and Ammar Khan and Amy Faranak and Ana Santos and Anthony Hevia and Antigona Unldreaj and Arash Aghagol and Arezoo Abdollahi and Aycha Tammour and Azadeh HajiHosseini and Bahareh Behroozi and Benjamin Ajibade and Bharat Saxena and Carlos Muñoz Ferrandis and Daniel McDuff and Danish Contractor and David Lansky and Davis David and Douwe Kiela and Duong A. Nguyen and Edward Tan and Emi Baylor and Ezinwanne Ozoani and Fatima Mirza and Frankline Ononiwu and Habib Rezanejad and Hessie Jones and Indrani Bhattacharya and Irene Solaiman and Irina Sedenko and Isar Nejadgholi and Jesse Passmore and Josh Seltzer and Julio Bonis Sanz and Livia Dutra and Mairon Samagaio and Maraim Elbadri and Margot Mieskes and Marissa Gerchick and Martha Akinlolu and Michael McKenna and Mike Qiu and Muhammed Ghauri and Mykola Burynok and Nafis Abrar and Nazneen Rajani and Nour Elkott and Nour Fahmy and Olanrewaju Samuel and Ran An and Rasmus Kromann and Ryan Hao and Samira Alizadeh and Sarmad Shubber and Silas Wang and Sourav Roy and Sylvain Viguier and Thanh Le and Tobi Oyebade and Trieu Le and Yoyo Yang and Zach Nguyen and Abhinav Ramesh Kashyap and Alfredo Palasciano and Alison Callahan and Anima Shukla and Antonio Miranda-Escalada and Ayush Singh and Benjamin Beilharz and Bo Wang and Caio Brito and Chenxi Zhou and Chirag Jain and Chuxin Xu and Clémentine Fourrier and Daniel León Periñán and Daniel Molano and Dian Yu and Enrique Manjavacas and Fabio Barth and Florian Fuhrimann and Gabriel Altay and Giyaseddin Bayrak and Gully Burns and Helena U. Vrabec and Imane Bello and Ishani Dash and Jihyun Kang and John Giorgi and Jonas Golde and Jose David Posada and Karthik Rangasai Sivaraman and Lokesh Bulchandani and Lu Liu and Luisa Shinzato and Madeleine Hahn de Bykhovetz and Maiko Takeuchi and Marc Pàmies and Maria A Castillo and Marianna Nezhurina and Mario Sänger and Matthias Samwald and Michael Cullan and Michael Weinberg and Michiel De Wolf and Mina Mihaljcic and Minna Liu and Moritz Freidank and Myungsun Kang and Natasha Seelam and Nathan Dahlberg and Nicholas Michio Broad and Nikolaus Muellner and Pascale Fung and Patrick Haller and Ramya Chandrasekhar and Renata Eisenberg and Robert Martin and Rodrigo Canalli and Rosaline Su and Ruisi Su and Samuel Cahyawijaya and Samuele Garda and Shlok S Deshmukh and Shubhanshu Mishra and Sid Kiblawi and Simon Ott and Sinee Sang-aroonsiri and Srishti Kumar and Stefan Schweter and Sushil Bharati and Tanmay Laud and Théo Gigant and Tomoya Kainuma and Wojciech Kusa and Yanis Labrak and Yash Shailesh Bajaj and Yash Venkatraman and Yifan Xu and Yingxin Xu and Yu Xu and Zhe Tan and Zhongli Xie and Zifan Ye and Mathilde Bras and Younes Belkada and Thomas Wolf},
      year={2023},
      eprint={2211.05100},
      archivePrefix={arXiv},
      primaryClass={cs.CL}
}

@article{multiberts2021,
  author    = {Thibault Sellam and
               Steve Yadlowsky and
               Jason Wei and
               Naomi Saphra and
               Alexander D'Amour and
               Tal Linzen and
               Jasmijn Bastings and
               Iulia Turc and
               Jacob Eisenstein and
               Dipanjan Das and
               Ian Tenney and
               Ellie Pavlick},
  title     = {The MultiBERTs: {BERT} Reproductions for Robustness Analysis},
  journal   = {CoRR},
  volume    = {abs/2106.16163},
  year      = {2021},
  url       = {https://arxiv.org/abs/2106.16163},
  eprinttype = {arXiv},
  eprint    = {2106.16163},
  bibsource = {dblp computer science bibliography, https://dblp.org}
}

@misc{brown2020gpt3,
      title={Language Models are Few-Shot Learners}, 
      author={Tom B. Brown and Benjamin Mann and Nick Ryder and Melanie Subbiah and Jared Kaplan and Prafulla Dhariwal and Arvind Neelakantan and Pranav Shyam and Girish Sastry and Amanda Askell and Sandhini Agarwal and Ariel Herbert-Voss and Gretchen Krueger and Tom Henighan and Rewon Child and Aditya Ramesh and Daniel M. Ziegler and Jeffrey Wu and Clemens Winter and Christopher Hesse and Mark Chen and Eric Sigler and Mateusz Litwin and Scott Gray and Benjamin Chess and Jack Clark and Christopher Berner and Sam McCandlish and Alec Radford and Ilya Sutskever and Dario Amodei},
      year={2020},
      eprint={2005.14165},
      archivePrefix={arXiv},
      primaryClass={cs.CL},
      url={https://arxiv.org/abs/2005.14165}, 
}

@inproceedings{gemini2023,
author = {Wang, Zhuang and Jia, Zhen and Zheng, Shuai and Zhang, Zhen and Fu, Xinwei and Ng, T. S. Eugene and Wang, Yida},
title = {GEMINI: Fast Failure Recovery in Distributed Training with In-Memory Checkpoints},
year = {2023},
isbn = {9798400702297},
publisher = {Association for Computing Machinery},
address = {New York, NY, USA},
url = {https://doi.org/10.1145/3600006.3613145},
doi = {10.1145/3600006.3613145},
booktitle = {Proceedings of the 29th Symposium on Operating Systems Principles},
pages = {364–381},
numpages = {18},
location = {Koblenz, Germany},
series = {SOSP '23}
}

@misc{maeng2020cprunderstandingimprovingfailure,
      title={CPR: Understanding and Improving Failure Tolerant Training for Deep Learning Recommendation with Partial Recovery}, 
      author={Kiwan Maeng and Shivam Bharuka and Isabel Gao and Mark C. Jeffrey and Vikram Saraph and Bor-Yiing Su and Caroline Trippel and Jiyan Yang and Mike Rabbat and Brandon Lucia and Carole-Jean Wu},
      year={2020},
      eprint={2011.02999},
      archivePrefix={arXiv},
      primaryClass={cs.LG},
      url={https://arxiv.org/abs/2011.02999}, 
}

@inproceedings {checkfreq,
author = {Jayashree Mohan and Amar Phanishayee and Vijay Chidambaram},
title = {{CheckFreq}: Frequent, {Fine-Grained} {DNN} Checkpointing},
booktitle = {19th USENIX Conference on File and Storage Technologies (FAST 21)},
year = {2021},
isbn = {978-1-939133-20-5},
pages = {203--216},
url = {https://www.usenix.org/conference/fast21/presentation/mohan},
publisher = {USENIX Association},
month = feb
}

@inproceedings {checknrun,
author = {Assaf Eisenman and Kiran Kumar Matam and Steven Ingram and Dheevatsa Mudigere and Raghuraman Krishnamoorthi and Krishnakumar Nair and Misha Smelyanskiy and Murali Annavaram},
title = {{Check-N-Run}: a Checkpointing System for Training Deep Learning Recommendation Models},
booktitle = {19th USENIX Symposium on Networked Systems Design and Implementation (NSDI 22)},
year = {2022},
isbn = {978-1-939133-27-4},
address = {Renton, WA},
pages = {929--943},
url = {https://www.usenix.org/conference/nsdi22/presentation/eisenman},
publisher = {USENIX Association},
month = apr
}

@inproceedings{failures2017,
author = {Gupta, Saurabh and Patel, Tirthak and Engelmann, Christian and Tiwari, Devesh},
title = {Failures in large scale systems: long-term measurement, analysis, and implications},
year = {2017},
isbn = {9781450351140},
publisher = {Association for Computing Machinery},
address = {New York, NY, USA},
url = {https://doi.org/10.1145/3126908.3126937},
doi = {10.1145/3126908.3126937},
booktitle = {Proceedings of the International Conference for High Performance Computing, Networking, Storage and Analysis},
articleno = {44},
numpages = {12},
location = {Denver, Colorado},
series = {SC '17}
}

@INPROCEEDINGS{analysisfailure2014,
  author={Di Martino, Catello and Kalbarczyk, Zbigniew and Iyer, Ravishankar K. and Baccanico, Fabio and Fullop, Joseph and Kramer, William},
  booktitle={2014 44th Annual IEEE/IFIP International Conference on Dependable Systems and Networks}, 
  title={Lessons Learned from the Analysis of System Failures at Petascale: The Case of Blue Waters}, 
  year={2014},
  volume={},
  number={},
  pages={610-621},
  doi={10.1109/DSN.2014.62}}

@book{fundamentals2017,
author = {Buduma, Nikhil and Locascio, Nicholas},
title = {Fundamentals of Deep Learning: Designing Next-Generation Machine Intelligence Algorithms},
year = {2017},
isbn = {1491925612},
publisher = {O'Reilly Media, Inc.},
edition = {1st}
}

@INPROCEEDINGS{bfloat2019,
  author={Burgess, Neil and Milanovic, Jelena and Stephens, Nigel and Monachopoulos, Konstantinos and Mansell, David},
  booktitle={2019 IEEE 26th Symposium on Computer Arithmetic (ARITH)}, 
  title={Bfloat16 Processing for Neural Networks}, 
  year={2019},
  volume={},
  number={},
  pages={88-91},
  doi={10.1109/ARITH.2019.00022}}

@TechReport{BWT,
    author = {Michael Burrows and David J. Wheeler},
    title = {A Block-Sorting Lossless Data Compression Algorithm},
    institution={Digital Systems Research Center},
    year="1994",
    month="May",
    number="124" 
}

@misc{rfc1951,
    series =    {Request for Comments},
    number =    1951,
    howpublished =  {RFC 1951},
    publisher = {RFC Editor},
    doi =       {10.17487/RFC1951},
    url =       {https://www.rfc-editor.org/info/rfc1951},
    author =    {L. Peter Deutsch},
    title =     {{DEFLATE Compressed Data Format Specification version 1.3}},
    pagetotal = 17,
    year =      1996,
    month =     may,
}

@article{Hirsch1990,
  title={Efficient Decoding of Prefix Codes},
  author={Daniel S. Hirschberg and Debra A. Lelewer},
  journal={Communications of the ACM},
  volume={33},
  pages={449--459},
  year={1990},
  publisher={ACM}
}

@misc{park2024comprehensivesurveycompressionalgorithms,
      title={A Comprehensive Survey of Compression Algorithms for Language Models}, 
      author={Seungcheol Park and Jaehyeon Choi and Sojin Lee and U Kang},
      year={2024},
      eprint={2401.15347},
      archivePrefix={arXiv},
      primaryClass={cs.CL},
      url={https://arxiv.org/abs/2401.15347}, 
}

@article{10.14778/3648160.3648180,
author = {Chen, Xinyu and Tian, Jiannan and Beaver, Ian and Freeman, Cynthia and Yan, Yan and Wang, Jianguo and Tao, Dingwen},
title = {FCBench: Cross-Domain Benchmarking of Lossless Compression for Floating-Point Data},
year = {2024},
issue_date = {February 2024},
publisher = {VLDB Endowment},
volume = {17},
number = {6},
issn = {2150-8097},
url = {https://doi.org/10.14778/3648160.3648180},
doi = {10.14778/3648160.3648180},
journal = {Proc. VLDB Endow.},
month = {may},
pages = {1418–1431},
numpages = {14}
}

@misc{hershcovitch2024losslessnearlosslesscompressionfoundation,
      title={Lossless and Near-Lossless Compression for Foundation Models}, 
      author={Moshik Hershcovitch and Leshem Choshen and Andrew Wood and Ilias Enmouri and Peter Chin and Swaminathan Sundararaman and Danny Harnik},
      year={2024},
      eprint={2404.15198},
      archivePrefix={arXiv},
      primaryClass={cs.LG},
      url={https://arxiv.org/abs/2404.15198}, 
}

@INPROCEEDINGS{8546121,
  author={Mogahed, Hussein Sh. and Yakunin, Alexey G.},
  booktitle={2018 XIV International Scientific-Technical Conference on Actual Problems of Electronics Instrument Engineering (APEIE)}, 
  title={Development of a Lossless Data Compression Algorithm for Multichannel Environmental Monitoring Systems}, 
  year={2018},
  volume={},
  number={},
  pages={483-486},
  doi={10.1109/APEIE.2018.8546121}}

@article{10.1145/3560814,
author = {Chiarot, Giacomo and Silvestri, Claudio},
title = {Time Series Compression Survey},
year = {2023},
issue_date = {October 2023},
publisher = {Association for Computing Machinery},
address = {New York, NY, USA},
volume = {55},
number = {10},
issn = {0360-0300},
url = {https://doi.org/10.1145/3560814},
doi = {10.1145/3560814},
journal = {ACM Comput. Surv.},
month = {feb},
articleno = {198},
numpages = {32}
}

\appendix
\onecolumn
\newpage
\section*{Appendix A}

\subsection*{A.1  Data Set Descriptions}
\label{appendix_a1}

\begin{itemize}
    \item \textbf{Bloom} - BigScience Large Open-science Open-access Multilingual Language Model (BLOOM) \cite{workshop2023bloom} \\
    https://huggingface.co/bigscience/bloom.
    \item \textbf{Amber} - Amber is a 7B English language model using the LLaMA architecture.\\https://huggingface.co/LLM360/Amber
    \item \textbf{Transnormer} - Linear transformer aimed at reducing the quadratic space-time complexity of vanilla transformers. TransNormerLLM is the first linear attention-based LLM that outperforms conventional softmax attention-based models in terms of both accuracy and efficiency. 
    \\ https://huggingface.co/OpenNLPLab/TransNormerLLM-7B
    \item \textbf{Olmo} - OLMo is a series of Open Language Models designed to enable the science of language models. The OLMo models are trained on the Dolma dataset.
    \\ https://huggingface.co/allenai/OLMo-7B
    \item \textbf{Multiberts} - MultiBERTs (pretrained BERT) model on English language using a masked language modeling (MLM) objective. The model is based on the paper by Sellam et al. ~\cite{multiberts2021}.
    \\  https://huggingface.co/MultiBertGunjanPatrick/multiberts-seed-0
    \item \textbf{Gpt2} - A reproduction of GPT-2 based on 124M parameters trained using x8 A100 GPUs on OpenWebText.
    \\ https://github.com/karpathy/nanoGPT
\end{itemize}

\begin{table*}[ht!]
\subsection*{A.2 Tensor Dimensions for Sample Data}
\label{appendix_a2}

    \centering
    \small
    \scalebox{0.9}{
    \begin{tabular}{ccccccc}
        \hline
         \textbf{Bloom}&  \textbf{Amber}&  \textbf{Transnormer}&    \textbf{Multiberts}&  \textbf{Olmo}& \textbf{GPT2}\\
         \hline
         14336 (425) &  32000 x 4096 (2) &  5120 (44) &  512 x 512 (1) &  50304 x 4096 (2)& 1024 x 768 (1)\\
         43008 x 14336 (70)&  4096 (65)&  5120 x 5120 (42) &  30522 x 768 (2) &  12288 x 4096 (32)& 50257 x 768 (1)\\
         43008 (70) &  4096 x 4096 (128) &  5120 x 15360 (42)&  512 x 768 (1) &  4096 x 4096 (32)& 1024 x 1024 (12)\\
         14336 x 14336 (70) &  11008 x 4096 (64) &  15360 x 5120 (126) &  1536 x 768 (2) &  4096 x 11008 (32)& 2304 (12)\\
         57344 x  4336 (70) &  4096 x 11008 (32) &  5120 x 15360 (42)&   768 (114) &  22016 x 4096 (32)& 768 x 2304 (12)\\
         57344 (70) &  &  &  &  768 x 768 (50) &  768 (74)\\
         \hline
    \end{tabular}
    }
    \caption{Tensor shard shapes for analyzed data sets (count in parenthesis)}
    \label{tab:shapes}
\end{table*}


\begin{table*}
\section*{A.3  Detailed Results for BLOOM Data Set}
\vspace{3mm}
\begin{tabularx}{0.9\linewidth}{
>{\setlength{\hsize}{.1\hsize}\raggedright\footnotesize}X
>{\setlength{\hsize}{.4\hsize}\raggedright\footnotesize}X
>{\setlength{\hsize}{.1\hsize}\raggedright\footnotesize}X
>{\setlength{\hsize}{.1\hsize}\raggedright\footnotesize}X
>{\setlength{\hsize}{.1\hsize}\raggedright\footnotesize}X
>{\setlength{\hsize}{.1\hsize}\raggedright\footnotesize}X
>{\setlength{\hsize}{.1\hsize}\raggedright\arraybackslash\footnotesize}X }
\hline
\textbf{Encoder} & \textbf{Shard} & \textbf{Mean Compression Ratio} & \textbf{Shard Size} & \textbf{Mean Encode Time (seconds)} & \textbf{Mean Decode Time (seconds)} & \textbf{Steps} \\
\hline

         BG-LMC   & h.X.mlp.dense\_4h\_to\_h.weight                       & 0.565 & 1.5GiB   & 4.6762 & 4.5397 & 200 \\
BG-LMC   & h.X.mlp.dense\_h\_to\_4h.weight                       & 0.530 & 1.5GiB   & 4.7379 & 4.6953 & 200 \\
BG-LMC   & h.X.self\_attention.dense.weight                    & 0.600 & 392.0MiB & 1.0748 & 1.0068 & 200 \\
BG-LMC   & h.X.self\_attention.query\_key\_value.weight          & 0.519 & 1.1GiB   & 3.4900 & 3.3678 & 200 \\
BG-LMC   & word\_embeddings.weight                             & 0.574 & 6.7GiB   & 23.5087 & 24.0144 & 20 \\
\hline
LMC      & h.X.mlp.dense\_4h\_to\_h.weight                       & 0.647 & 1.5GiB   & 3.6747 & 3.1807 & 200 \\
LMC      & h.X.mlp.dense\_h\_to\_4h.weight                       & 0.603 & 1.5GiB   & 3.6340 & 3.2266 & 200 \\
LMC      & h.X.self\_attention.dense.weight                    & 0.695 & 392.0MiB & 0.9288 & 0.8136 & 200 \\
LMC      & h.X.self\_attention.query\_key\_value.weight          & 0.594 & 1.1GiB   & 2.7130 & 2.3578 & 200 \\
LMC      & word\_embeddings.weight                             & 0.664 & 6.7GiB   & 17.0775 & 16.1114 & 20 \\
\hline
BG-BZ2   & h.X.mlp.dense\_4h\_to\_h.weight                       & 0.593 & 1.5GiB   & 154.3418 & 79.0762 & 200 \\
BG-BZ2   & h.X.mlp.dense\_h\_to\_4h.weight                       & 0.542 & 1.5GiB   & 137.3355 & 72.3389 & 200 \\
BG-BZ2   & h.X.self\_attention.dense.weight                    & 0.627 & 392.0MiB & 41.7933 & 20.6631 & 200 \\
BG-BZ2   & h.X.self\_attention.query\_key\_value.weight          & 0.538 & 1.1GiB   & 103.2792 & 54.1697 & 200 \\
BG-BZ2   & word\_embeddings.weight                             & 0.599 & 6.7GiB   & 673.7042 & 344.8112 & 20 \\
\hline
BG-LZ4   & h.X.mlp.dense\_4h\_to\_h.weight                       & 0.722 & 1.5GiB   & 4.3664 & 1.6828 & 200 \\
BG-LZ4   & h.X.mlp.dense\_h\_to\_4h.weight                       & 0.669 & 1.5GiB   & 4.0569 & 1.6928 & 200 \\
BG-LZ4   & h.X.self\_attention.dense.weight                    & 0.754 & 392.0MiB & 1.1768 & 0.4081 & 200 \\
BG-LZ4   & h.X.self\_attention.query\_key\_value.weight          & 0.670 & 1.1GiB   & 3.1123 & 1.3113 & 200 \\
BG-LZ4   & word\_embeddings.weight                             & 0.725 & 6.7GiB   & 21.4897 & 9.5665 & 20 \\
\hline
BZ2      & h.X.mlp.dense\_4h\_to\_h.weight                       & 0.593 & 1.5GiB   & 171.0602 & 85.7883 & 200 \\
BZ2      & h.X.mlp.dense\_h\_to\_4h.weight                       & 0.543 & 1.5GiB   & 157.7029 & 79.5812 & 200 \\
BZ2      & h.X.self\_attention.dense.weight                    & 0.627 & 392.0MiB & 43.2183 & 21.4598 & 200 \\
BZ2      & h.X.self\_attention.query\_key\_value.weight          & 0.535 & 1.1GiB   & 117.3340 & 60.5858 & 200 \\
BZ2      & word\_embeddings.weight                             & 0.598 & 6.7GiB   & 734.9330 & 371.4208 & 20 \\
\hline
LZ4      & h.X.mlp.dense\_4h\_to\_h.weight                       & 0.948 & 1.5GiB   & 3.2303 & 1.3582 & 200 \\
LZ4      & h.X.mlp.dense\_h\_to\_4h.weight                       & 0.888 & 1.5GiB   & 3.7749 & 1.5014 & 200 \\
LZ4      & h.X.self\_attention.dense.weight                    & 0.976 & 392.0MiB & 0.6223 & 0.3190 & 200 \\
LZ4      & h.X.self\_attention.query\_key\_value.weight          & 0.885 & 1.1GiB   & 2.8173 & 1.1237 & 200 \\
LZ4      & word\_embeddings.weight                             & 0.964 & 6.7GiB   & 14.3532 & 7.8359 & 20 \\
\hline
DEFLATE  & h.X.mlp.dense\_4h\_to\_h.weight                       & 0.679 & 1.5GiB   & 34.7788 & 18.3078 & 200 \\
DEFLATE  & h.X.mlp.dense\_h\_to\_4h.weight                       & 0.619 & 1.5GiB   & 35.0574 & 16.4662 & 200 \\
DEFLATE  & h.X.self\_attention.dense.weight                    & 0.722 & 392.0MiB & 8.2072 & 4.8629 & 200 \\
DEFLATE  & h.X.self\_attention.query\_key\_value.weight          & 0.616 & 1.1GiB   & 26.7587 & 12.4380 & 200 \\
DEFLATE  & word\_embeddings.weight                             & 0.691 & 6.7GiB   & 145.9071 & 81.3236 & 20 \\
\hline
gzip     & h.X.mlp.dense\_4h\_to\_h.weight                       & 0.679 & 1.5GiB   & 35.0080 & 18.5110 & 200 \\
gzip     & h.X.mlp.dense\_h\_to\_4h.weight                       & 0.619 & 1.5GiB   & 34.9759 & 16.5045 & 200 \\
gzip     & h.X.self\_attention.dense.weight                    & 0.722 & 392.0MiB & 8.1997 & 4.8827 & 200 \\
gzip     & h.X.self\_attention.query\_key\_value.weight          & 0.616 & 1.1GiB   & 26.7039 & 12.5117 & 200 \\
gzip     & word\_embeddings.weight                             & 0.691 & 6.7GiB   & 145.7747 & 82.2829 & 20 \\
\hline

    \end{tabularx}
    \label{tab:detailed_bloom}
\end{table*}

\end{document}